\documentclass{article}
\usepackage[preprint]{corl_2026}
\usepackage{amsmath}
\usepackage{amssymb}
\usepackage{graphicx}
\usepackage{array}
\usepackage{booktabs}
\usepackage{wrapfig}
\usepackage{float}
\usepackage[font=small]{caption}
\usepackage{etoc}
\newcommand{\method}{UniLab}
\newcommand{\affilmark}[1]{\textsuperscript{#1}}
\newcommand{\coremark}{\textsuperscript{*}}
\newcommand{\advisemark}{\textsuperscript{\ensuremath{\dagger}}}

\begin{document}

\setcounter{tocdepth}{-1}

\title{\method{}: A Heterogeneous Architecture for Robot RL Beyond GPU-Dominant Paradigms}



\author{
\normalfont
Yufei Jia\affilmark{1}\coremark,
Zhanxiang Cao\affilmark{2, 3}\coremark,
Mingrui Yu\affilmark{1}\coremark,
Heng Zhang\affilmark{4}\coremark,
Shenyu Chen\affilmark{5}\coremark,
Dixuan Jiang\affilmark{6}\coremark,\\
Meng Li\affilmark{7},
Xiaofan Li\affilmark{7},
Yiyang Liu\affilmark{1},
Junzhe Wu\affilmark{1},
Zheng Li\affilmark{11},
XiLin Fang\affilmark{8},\\
Ting-Yu Tsui\affilmark{1},
Shengcheng Fu\affilmark{9, 3},
Haoyang Li\affilmark{2, 3},
Anqi Wang\affilmark{10},
Zifan Wang\affilmark{11},
Dongjie Zhu\affilmark{1},\\
Chenyu Cao\affilmark{12},
Zhenbiao Huang\affilmark{13},
Ziang Zheng\affilmark{1},
Jie Lu\affilmark{14},
Xin Ma\affilmark{15},
Zhengyang Wei\affilmark{15},\\
Xiang Zhao\affilmark{4},
Tianyue Zhan\affilmark{2, 3},
Ye He\affilmark{16},
Yuxiang Chen\affilmark{17},
Yizhou Jiang\affilmark{1},
Yue Li\affilmark{10},\\
Haizhou Ge\affilmark{1},
Yuhang Dong\affilmark{18},
Fan Jia\affilmark{19},
Ziheng Zhang\affilmark{19},
Meng Zhang\affilmark{19},
Xiwa Deng\affilmark{4},\\
Zhixing Chen\affilmark{1},
Hanyang Shao\affilmark{10},
Chenxin Dong\affilmark{19},
Yixuan Li\affilmark{6},
Yizhi Chen\affilmark{9, 3},\\
Bokui Chen\affilmark{1},
Kaifeng Zhang\affilmark{20},
Hanqing Cui\affilmark{4},
Yusen Qin\affilmark{21},
Ruqi Huang\affilmark{1},\\
Lei Han\affilmark{10}\advisemark,
Tiancai Wang\affilmark{19}\advisemark,
Xiang Li\affilmark{1}\advisemark,
Yue Gao\affilmark{2, 3}\advisemark,
Guyue Zhou\affilmark{1}\advisemark\\[0.6em]
{\normalfont\scriptsize
\affilmark{1}THU,
\affilmark{2}SJTU,
\affilmark{3}SII,
\affilmark{4}Motphys,
\affilmark{5}HITSZ,
\affilmark{6}BIT,
\affilmark{7}NEU,
\affilmark{8}SUSTech,
\affilmark{9}TJU,
\affilmark{10}DISCOVER Robotics,
\affilmark{11}HKUST(GZ)},\\
{\normalfont\scriptsize
\affilmark{12}Galbot,
\affilmark{13}NUS,
\affilmark{14}WTU,
\affilmark{15}HBUT,
\affilmark{16}AMD,
\affilmark{17}NJU,
\affilmark{18}ZJU,
\affilmark{19}Dexmal,
\affilmark{20}Sharpa,
\affilmark{21}D-Robotics}\\
{\normalfont\scriptsize
\coremark Core contributors.
\advisemark Advising.
Correspondence to: Yufei Jia \texttt{<jyf23@mails.tsinghua.edu.cn>}.}
}

\maketitle

\begin{center}
  \vspace{-20pt}
  \begin{minipage}{0.98\textwidth}
    \centering
    \includegraphics[width=\linewidth]{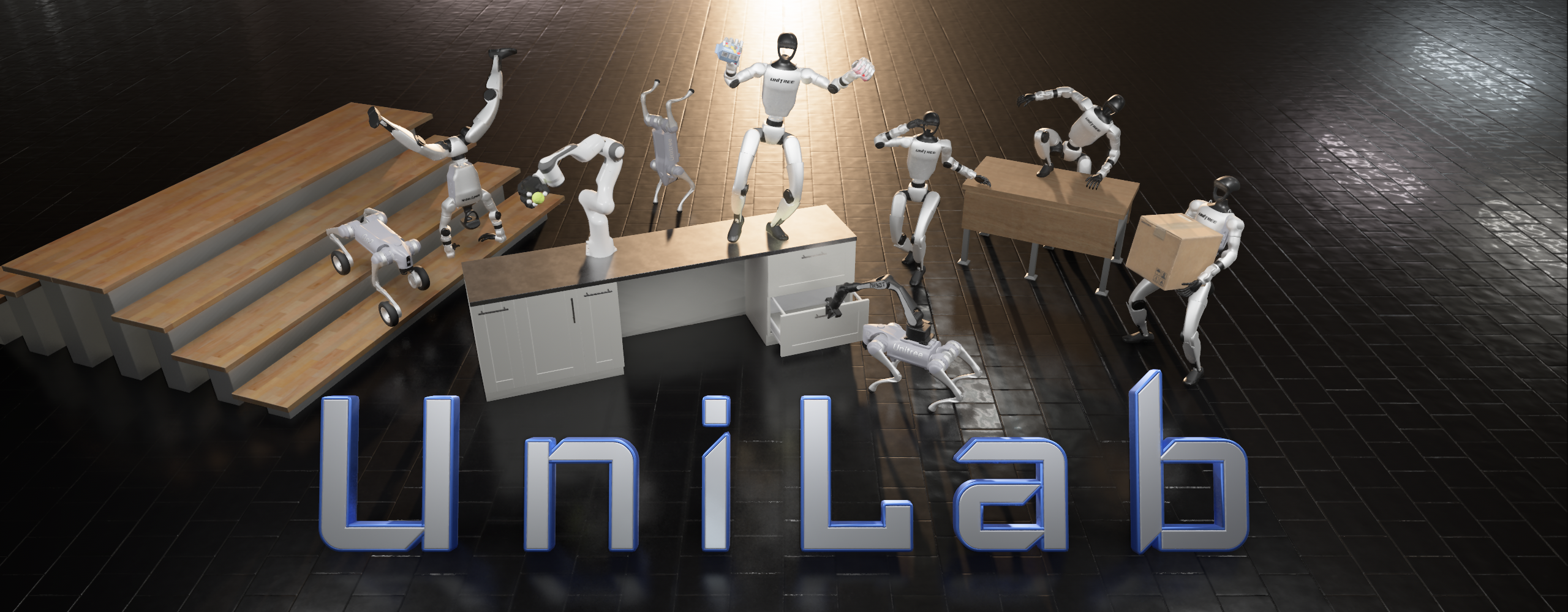}
    \ifdefined\captionsetup
      \captionsetup{hypcap=false}%
    \fi
    \makeatletter\def\@captype{figure}\makeatother
    \caption{Teaser. Representative robot-control tasks in \method{}; ``Uni'' means unified cross-platform training. Teaser image rendered with MotrixSim.}
    \label{fig:teaser}
  \end{minipage}
\end{center}

\begin{abstract}%

Simulation-based RL for contemporary robot control is increasingly organized around GPU-resident simulation: physics, rollout collection, and learning are placed on a single GPU-centric execution path. This paradigm has greatly improved training speed, but it has also encouraged a default assumption that efficient training requires physics to reside on the GPU. We revisit this assumption. Our view is that, in simulation-dominated robot control, the essential question is not which processor runs physics, but whether simulation throughput, policy learning, and runtime synchronization form an efficient end-to-end loop. We present \method{}, a heterogeneous CPU-simulation / GPU-learning architecture that decouples CPU-parallel simulation from GPU policy updates through a unified runtime for data movement, buffering, and synchronization. \method{} is implemented as a complete and extensible training system using MuJoCoUni and MotrixSim CPU-batched physics backends, supporting PPO, FastSAC, FlashSAC, and APPO. On representative simulation-based robot control tasks, \method{} improves end-to-end training efficiency by 3--10$\times$ under the same hardware configuration, while reducing dependence on the NVIDIA CUDA-based software stack and supporting cross-platform execution on the Apple macOS platform and the AMD ROCm and Intel XPU accelerator backends. These results show that GPU simulation is an effective path to efficient training, but not a necessary one, broadening the practical system choices available for robot RL training. 
Project page: \url{https://unilabsim.github.io}.

\end{abstract}

\section{Introduction}
\label{sec:introduction}

Training infrastructure has become a first-order factor in simulation-based robot RL: faster training reduces the wall-clock cost of a single experiment, shortens system and algorithm iteration cycles, and expands the range of tasks that can be studied under practical hardware budgets. The dominant answer in recent years has been clear: place physics simulation, rollout collection, and learning on a GPU-centric execution path; Isaac Gym, Isaac Lab, MuJoCo Playground, mjlab, ManiSkill3, and Genesis show that large-scale GPU-resident environment parallelism can greatly accelerate robot control training \citep{makoviychuk2021isaac, mittal2025isaac, zakka2025mujoco, Zakka_mjlab_A_Lightweight_2026, taomaniskill3, Genesis}. This success has shaped the current community default that efficient training should be organized around GPU-resident physics, tying high-throughput experimentation to a narrower set of GPU-resident software environments.

Robot RL training, however, is a closed-loop system coupling data generation, policy updates, and synchronization constraints, not a simulator benchmark alone. In simulation-dominated tasks, end-to-end efficiency depends on simulation throughput, learner utilization, collector--learner synchronization, data movement and buffering overhead, and whether hardware is allocated to the stage that actually limits wall-clock time: the learner may wait for rollouts, collectors may wait for new parameters, and data movement or buffering may erase parallel gains. Whether physics runs on the GPU is therefore one design choice within a broader systems organization problem.

High-throughput environment execution is also possible outside GPU-resident physics. General RL systems have long used CPU-side vectorized or batched environments, and robot RL has precedents for CPU-distributed or CPU-parallel simulation, including OpenAI's Rubik's-cube hand system and recent RaiSim-based locomotion work \citep{weng2022envpool, liang2021rllib, tianshou, suarez2025pufferlib, akkaya2019solving, kim2024not, pearce2019exploring}. Algorithmic data dependencies further shape this organization: PPO preserves the strongest rollout/update synchronization constraint; APPO allows collection and learning to overlap while remaining close to the on-policy setting; and off-policy methods such as FastSAC and FlashSAC further relax the dependence of each update on trajectories from the latest policy \citep{schulman2017proximal, haarnoja2018soft}. This ordering lets us study algorithms as synchronization regimes rather than as separate algorithmic contributions: PPO tests whether CPU simulation can sustain strictly synchronized training, APPO tests collector--learner overlap once synchronization is relaxed, and FastSAC/FlashSAC test the replay-based producer--consumer path. This motivates the systems question studied here: can CPU-side batched rigid-body simulation, GPU-side policy learning, and the runtime path between them form an efficient training loop?

This paper asks whether efficient simulation-based robot control training must rely on GPU-resident simulation. Our thesis is that simulation-dominated robot control training requires high-throughput, well-coordinated simulation-learning execution, rather than GPU-resident simulation itself. We focus on representative robot control tasks in simulation, leaving real-world RL and vision-dominated settings outside the scope of this paper.

We present \method{}, a heterogeneous CPU-simulation / GPU-learning training architecture. CPU-side MuJoCoUni~\citep{MuJoCoUni} and MotrixSim~\citep{motrixsim2026} backends perform batched rigid-body simulation and data generation, GPU resources perform policy and value learning, and a unified runtime coordinates data movement, buffering, and synchronization. \method{} is a training-system organization rather than a new policy optimization algorithm; it is implemented as a complete and extensible training system with unified training and evaluation entrypoints and explicit task/backend interfaces, while supporting PPO, FastSAC, FlashSAC, and APPO in one framework.

Across representative simulated robot-control benchmarks, \method{} improves end-to-end training efficiency by 3--10$\times$ on the same single-GPU/single-CPU workstation, while reducing dependence on the NVIDIA CUDA-based software stack and supporting execution on Apple macOS, AMD ROCm, and Intel XPU backends. Our contributions are threefold:
\begin{list}{}{\leftmargin=0pt \labelwidth=0pt \labelsep=0pt \itemindent=0pt \itemsep=0pt \topsep=1pt \parsep=0pt \partopsep=0pt}
    \item \textbf{Systems framing.} We recast efficient robot RL training as a systems organization problem for the simulation-learning closed loop, rather than a consequence of GPU-resident physics alone.
    \item \textbf{Heterogeneous training architecture.} We present \method{}, which connects CPU-batched physics backends, a GPU learner, data buffering, and parameter synchronization through a unified runtime, while supporting PPO, FastSAC, FlashSAC, and APPO in one framework.
    \item \textbf{End-to-end evidence.} We show 3--10$\times$ wall-clock gains across robot embodiments, control workloads, and practical algorithms, together with execution evidence on macOS, ROCm, and XPU backends.
\end{list}

\section{Related Work}
\label{sec:related_work}

\subsection{GPU-resident robot learning.}
\begin{wraptable}{r}{0.50\textwidth}
    \vspace{-33pt}
    \centering
    \caption{Representative robot RL training systems.}
    \label{tab:related_systems_comparison}
    \scriptsize
    \setlength{\tabcolsep}{2.2pt}
    \renewcommand{\arraystretch}{1.08}
    \newcommand{\tablehead}[1]{\begin{tabular}[c]{@{}c@{}}\bfseries #1\end{tabular}}
    \newcommand{\tablethickline}{\noalign{\hrule height 0.6pt}}
    \begin{tabular*}{\linewidth}{@{\extracolsep{\fill}}cccc@{}}
        \tablethickline
        \rule[-0.9ex]{0pt}{4.5ex}\tablehead{System} & \tablehead{Phys.} & \tablehead{Batch} & \tablehead{Coupling} \\
        \noalign{\vskip 1pt}
        \hline
        \texttt{IsaacGym}  & PhysX   & GPU-C     & GPU-sync \\
        \texttt{IsaacLab}  & PhysX   & GPU-C     & GPU-sync \\
        \texttt{Genesis}   & Taichi  & GPU-C/M/R & GPU-sync \\
        \texttt{MJP}       & MJX     & GPU-C     & GPU-sync \\
        \texttt{MjLab}     & MJWarp  & GPU-C     & GPU-sync \\
        \hline
        \method{}          & MJU/Mtx & CPU       & H-async/sync \\
        \tablethickline
    \end{tabular*}
    \vspace{-6pt}
    \begin{minipage}{\linewidth}
        \scriptsize
        \raggedright
        \vspace{2pt}
        \emph{Note.} GPU-C/M/R: GPU batched physics on CUDA/Metal/ROCm. GPU-sync: synchronized GPU simulation--learning; H-async/sync: CPU simulation with GPU learning. MJU/Mtx/MJP: MuJoCoUni/MotrixSim/MuJoCo\_playground.
    \end{minipage}
    \vspace{-6pt}
\end{wraptable}

The dominant systems path for efficient robot RL training has been to place physics simulation, rollout collection, and learning on a GPU-centric execution path \citep{makoviychuk2021isaac, freeman2021brax, liang2018gpu}. MuJoCo provides a widely used foundation for robot control simulation \citep{todorov2012mujoco}, while Isaac Gym, Isaac Lab, MuJoCo Playground, mjlab, ManiSkill3, and Genesis have made large-scale GPU-resident environment parallelism a standard practice for robot learning \citep{makoviychuk2021isaac, mittal2025isaac, zakka2025mujoco, Zakka_mjlab_A_Lightweight_2026, taomaniskill3, Genesis}. Table~\ref{tab:related_systems_comparison} summarizes these systems along the axes most relevant to this paper: physics execution path, simulation--learning organization, and algorithmic data dependency.

\subsection{Systems lesson from GPU simulation.}
The central lesson from GPU-resident systems is the integration of fast physics execution with tightly coupled rollout collection and learner updates. For on-policy methods such as PPO, this organization fits synchronized batched rollout/update cycles and has proven effective across robot-control workloads \citep{schulman2017proximal, hwangbo2019learning, margolis2023walk, margolis2024rapid, wang2024arm, he2025asap, cao2026hiwet, bharthulwar2026staggered, shahid2024scaling}. We adopt this systems lesson but separate the training-system principle from one hardware path: efficient training requires low-overhead data generation, learning, and synchronization, while GPU kernels are most effective for regular, dense, and statically shaped execution; dynamic active contact sets, sparse interactions, collision handling, contact solving, closed-chain or other constraint handling, and contact-rich manipulation all stress this execution model.

\subsection{CPU-parallel environment execution.}
High-throughput environment execution also has a history outside GPU-resident physics. In general RL, EnvPool, RLlib, Tianshou, and PufferLib use CPU-side vectorized, batched, or parallel rollout collection as core system components \citep{weng2022envpool, liang2021rllib, tianshou, suarez2025pufferlib}. Robot RL also has CPU-distributed or CPU-parallel precedents, including OpenAI's Rubik's-cube hand system and recent RaiSim-based locomotion work \citep{akkaya2019solving, kim2024not}. These examples show that CPU-side environment parallelism is viable; \method{} asks whether, under the same hardware setting, modern CPU-batched simulation and a GPU learner can form an efficient end-to-end training path through a low-overhead runtime rather than only at extreme worker-cluster scale.

\subsection{Replay-based robot-control acceleration.}
Algorithmic data dependencies further shape the system organization. PPO is the practical default in many large-scale robot-training workloads, but its on-policy updates preserve strong synchronization between rollout generation and learner updates. Replay-based methods such as SAC and TD3 can reuse past experience and relax this dependence, while FastTD3, FastSAC, and FlashSAC show that this direction can accelerate high-dimensional robot control \citep{haarnoja2018soft, fujimoto2018addressing, seo2025fasttd3, seo2025learning, kim2026flashsac}. \method{} studies the complementary systems question: when data dependencies are relaxed, how can CPU simulation and GPU learning be coordinated to improve end-to-end wall-clock efficiency?

\section{\method{} Architecture}
\label{sec:architecture}

\begin{figure*}[t]
    \centering
    \includegraphics[width=\textwidth]{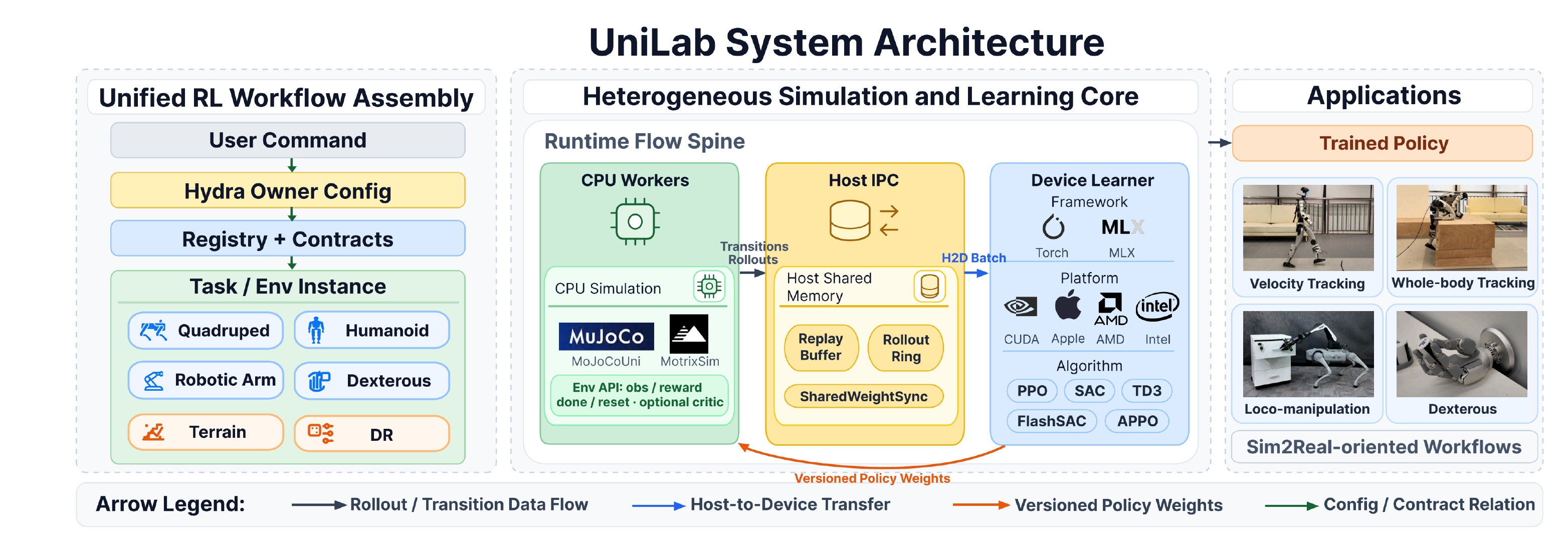}
    \caption{\method{} system architecture. The figure shows the data, scheduling, and parameter-synchronization paths between CPU-side batched physics backends, the unified runtime, and the GPU learner.}
    \label{fig:architecture_overview}
    \vspace{-15pt}
\end{figure*}

This section describes \method{} as an end-to-end training loop that combines CPU-side batched rigid-body simulation, GPU-side policy and value learning, and a unified runtime for coordinating the data path between them.

\subsection{Design Objective and Requirements}
\label{subsec:architecture_requirements}

The design objective is to improve the efficiency of the full simulation-learning loop without requiring GPU-resident simulation. \method{} follows hardware roles: CPUs generate large-scale simulation data, GPUs perform dense learning updates, and the runtime minimizes coordination cost. This objective induces three requirements.

\textbf{CPU-side simulation throughput.}
CPU-side batched rigid-body simulation must sustain enough throughput to continuously generate data for the workloads studied here.

\textbf{Non-blocking GPU learning.}
The GPU learner should consume buffered experience rather than idling behind rollout generation.

\textbf{Controlled runtime overhead.}
Data movement, buffering, and parameter synchronization must remain low-overhead so that the heterogeneous split does not degenerate into blocking handoffs.

\subsection{\method{} Execution Architecture}
\label{subsec:execution_architecture}

Figure~\ref{fig:architecture_overview} summarizes the system organization: \textbf{CPU workers} generate trajectories or transitions, the \textbf{GPU learner} performs policy and value updates, and the \textbf{unified runtime} coordinates data movement, buffering, scheduling, and parameter synchronization.

\begin{wrapfigure}[11]{r}{0.50\textwidth}
    \vspace{-6pt}
    \centering
    \includegraphics[width=\linewidth]{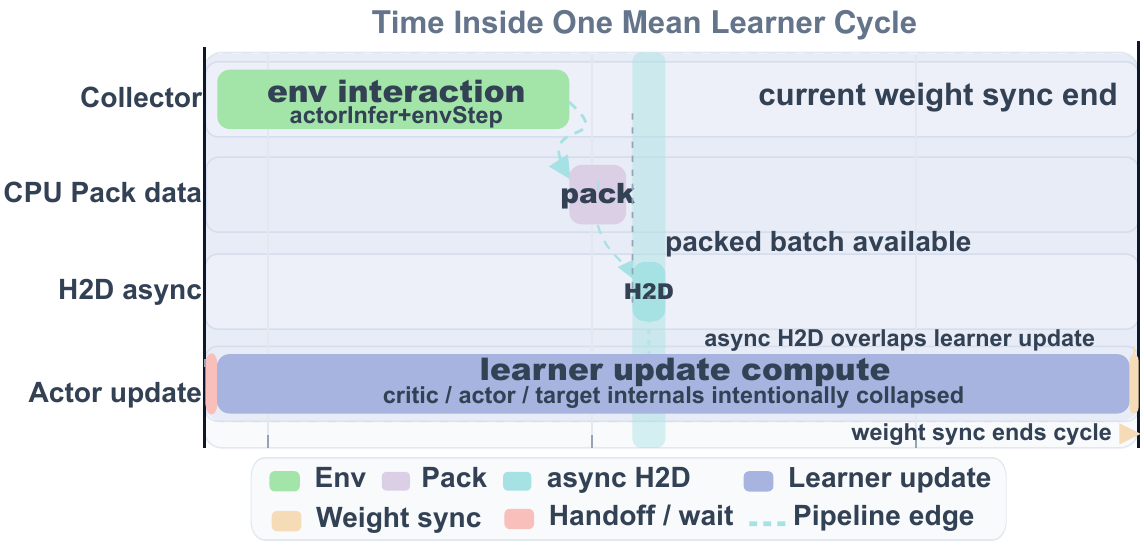}
    \caption{Collection--update timing and overlap.}
    \label{fig:system_attribution}
\end{wrapfigure}

\textbf{Collection--update timing and overlap.}
\method{} supports both synchronized and loosely coupled collection--update timing. Standard PPO uses a synchronized rollout/update cycle. Our APPO implementation follows the asynchronous on-policy formulation described by \citet{luo2019impact}: the collector writes fixed-horizon rollouts, behavior-policy log probabilities, and bootstrap information into a shared ring buffer while continuing to step the next rollout on the CPU; the learner drains available rollouts and performs V-trace correction and PPO-style updates on the GPU, with the V-trace clipping values listed in Appendix~\ref{app:algorithm_appo}. CPU collection and GPU learning therefore overlap in wall-clock time with parameter synchronization near rollout boundaries. FastSAC and FlashSAC use replay-based timing: collectors insert transition batches into a shared replay buffer, while the learner performs multiple updates from device batches; both variants use the same optimized runtime path, which requests CPU replay packing and device transfer for the next batch one tick ahead so they overlap with current learner updates.

Figure~\ref{fig:system_attribution} shows the FastSAC case, where collector-side work is staged ahead of learner computation and the main visible synchronization point is actor-weight handoff.

\textbf{Runtime abstraction.}
The unified runtime lets synchronized and loosely coupled execution share one system stack, connecting robot assets, task configurations, simulation backends, and learning algorithms through explicit interfaces.

\subsection{CPU Physics Backends and Task Interface}
\label{subsec:cpu_backends_task_interface}

\textbf{Batched CPU physics.}
\method{} realizes CPU-side throughput through backend-native batched environment execution: CPU workers advance environments at batch granularity and generate trajectories or transitions for the downstream learner.

\textbf{Backend contract.}
The current system connects two practical CPU-side simulation backends under a shared runtime contract. MuJoCoUni provides a CPU-batched MuJoCo runtime backend~\citep{MuJoCoUni}; the MotrixSim backend maps the same task and runtime contract onto the MotrixSim physics and rendering stack \citep{motrixsim2026}.

\textbf{Task and randomization interface.}
This contract covers task state, actions, observation-related data, reset and interval randomization hooks, terrain context, and playback capabilities, allowing physical parameters, observation perturbations, and task-condition changes to be scheduled by the training system rather than scattered across task scripts.

This design separates physics semantics, determined by the backend model and solver, from training throughput, determined by batched execution, data movement, and runtime coordination; the same learner binding can also target macOS, ROCm, and XPU, with backend-dependent throughput evaluated in Section~\ref{subsec:cross_platform}.

\section{Experiments}
\label{sec:experiments}

We evaluate three questions: whether CPU simulation provides enough throughput, whether heterogeneous CPU-simulation / GPU-learning improves end-to-end wall-clock efficiency, and whether the result is robust across task families and algorithms. The primary metric is end-to-end training efficiency; throughput measurements explain the mechanism.

\begin{figure*}[t]
    \centering
    \includegraphics[width=\textwidth]{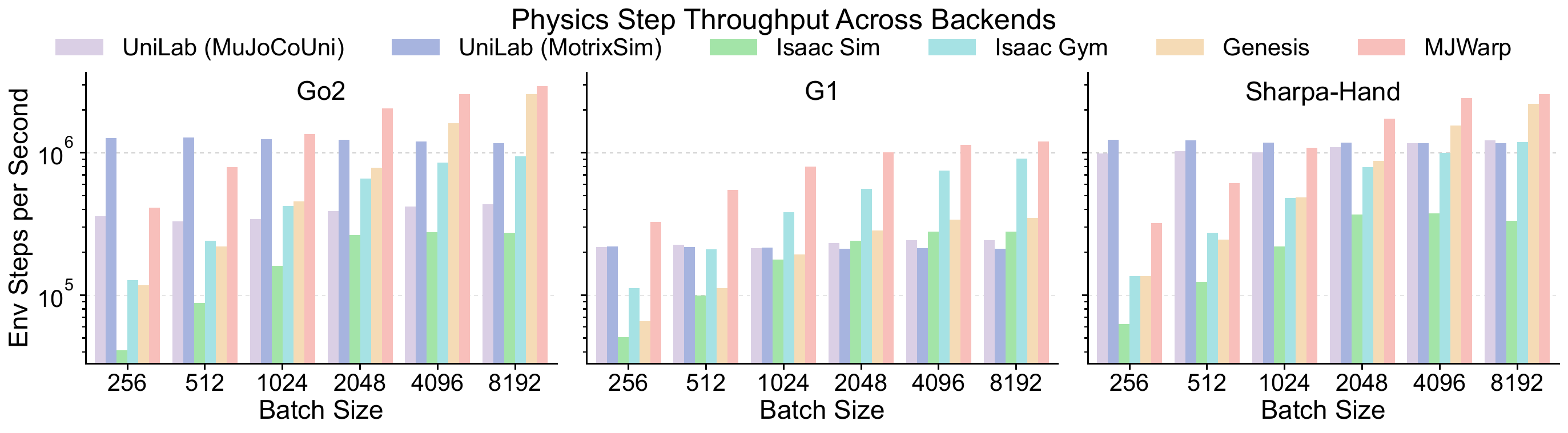}
    \caption{CPU simulation throughput across representative robot control scenes. The figure establishes the simulator-side capacity that underlies the end-to-end training results.}
    \vspace{-12pt}
    \label{fig:cpu_batch_throughput}
\end{figure*}

\subsection{Experimental Setup}
\label{subsec:experimental_setup}


Controlled comparisons use the same default Linux hardware: one NVIDIA RTX 4090 GPU, one AMD Ryzen 9 9950X3D CPU, and 64 GB of 4800 MT/s memory. Unless otherwise stated, \method{} results in the main experiments use the MuJoCoUni backend, while Apple macOS, AMD ROCm, and Intel XPU results are included as portability evidence. The task set spans locomotion, motion tracking, manipulation, and manipulation-locomotion across quadruped, wheeled-quadruped, humanoid, and dexterous-hand / in-hand manipulation embodiments. We organize algorithms by their synchronization constraints: PPO is the strictly synchronized on-policy baseline, APPO is the near-on-policy case where rollout collection can overlap with learning, and FastSAC/FlashSAC provides replay-based producer--consumer off-policy evidence. For comparisons against external baselines, we use their public task-resolved configurations and align controllable factors including observation spaces, action spaces, rewards, sensor noise, and the main domain-randomization settings, while preserving each system's native execution details. The reported results therefore reflect practical system-level wall-clock performance under the same hardware setting on representative task configurations. Detailed experimental setup is provided in Appendix~\ref{app:training_curves}.

\subsection{Can CPU Simulation Provide Enough Throughput for Robot RL?}
\label{subsec:cpu_sim_throughput}

In common robot-RL training settings, CPU physics does not necessarily provide lower throughput than GPU-based simulation; its relative advantage is more pronounced in workloads with complex contact and dexterous manipulation. Figure~\ref{fig:cpu_batch_throughput} and Table~\ref{tab:cpu_backend_task_throughput} show that batched CPU simulation provides the simulator-side capacity required by the heterogeneous execution model over the environment counts studied here.

\begin{wraptable}{r}{0.58\textwidth}
    \vspace{-20pt}
    \centering
    \caption{CPU env-step throughput ($10^3$ steps/s) by task and chip.}
    \label{tab:cpu_backend_task_throughput}
    \scriptsize
    \setlength{\tabcolsep}{1.4pt}
    \renewcommand{\arraystretch}{1.03}
    \begin{tabular*}{\linewidth}{@{\extracolsep{\fill}}lrrrrrr@{}}
        \toprule
        & \multicolumn{2}{c}{Go2} & \multicolumn{2}{c}{G1} & \multicolumn{2}{c}{Hand} \\
        \cmidrule(lr){2-3}\cmidrule(lr){4-5}\cmidrule(l){6-7}
        Chip & MJ & Motrix & MJ & Motrix & MJ & Motrix \\
        \midrule
        A18 Pro & 55.7 & 122.9 & 28.4 & 18.1 & 183.9 & 134.1 \\
        M5 Max & 288.0 & 797.8 & 178.8 & 127.7 & 1118.4 & 982.9 \\
        R9-8945HX & 246.2 & 704.2 & 154.6 & 113.6 & 434.1 & 542.2 \\
        TR-9980X & 915.9 & 2662.7 & 517.9 & 410.4 & 1991.5 & 2622.6 \\
        i7-11800H & 82.1 & 162.0 & 34.7 & 23.8 & 176.8 & 151.6 \\
        Xeon 8558 & 1002.4 & 847.2 & 424.6 & 379.5 & 2566.3 & 397.7 \\
        \bottomrule
    \end{tabular*}
    \vspace{-5pt}
    \begin{minipage}{\linewidth}
        \scriptsize
        \raggedright
        \vspace{2pt}
        \emph{Note.} Values are $10^3$ env steps/s; MJ = MuJoCoUni backend.
    \end{minipage}
    \vspace{-10pt}
\end{wraptable}

End-to-end training gives complementary evidence. In Figure~\ref{fig:end_to_end_training}(a), GPU-resident MjLab and CPU-step / GPU-learner \method{} achieve comparable efficiency on the same PPO task (Time Usage: 120.5/111.4 min vs. 109.3/109.2 min). Since synchronized PPO leaves little opportunity to hide rollout latency, PPO serves here as the strict synchronization stress test: this result indicates that CPU simulation is not the dominant bottleneck for this workload.

\subsection{Can CPU-Sim / GPU-Learn Improve End-to-End Efficiency?}
\label{subsec:end_to_end_training}

Given sufficient CPU-side throughput for strictly synchronized PPO, the next question is whether heterogeneous organization translates into end-to-end gains as data dependencies become looser. APPO remains near the on-policy regime but overlaps rollout collection with learning through correction; FastSAC/FlashSAC organize data generation as replay-based producer--consumer paths. Once the runtime decouples the learner from the collector, these more loosely coupled settings obtain 3--10$\times$ improvements in end-to-end training efficiency across multiple robot control tasks.

\begin{figure*}[t]
    \centering
    \includegraphics[width=\textwidth]{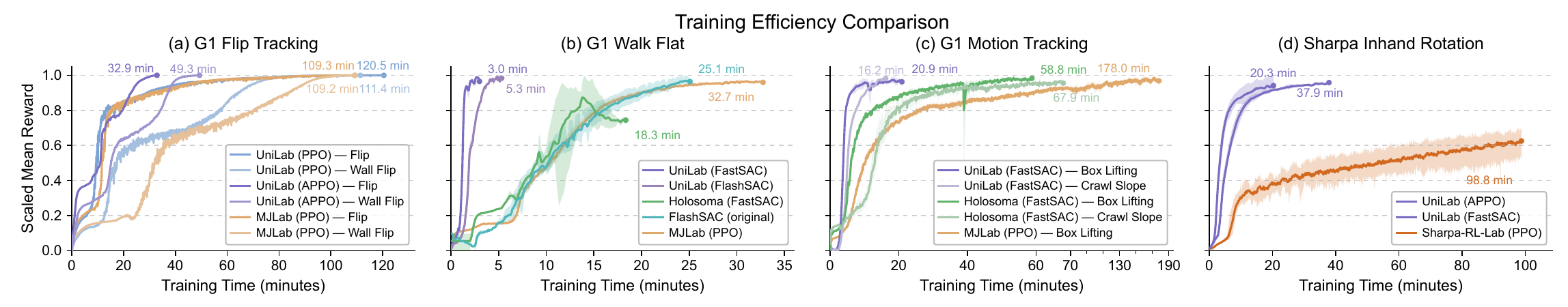}
    \vspace{-20pt}
    \caption{End-to-end training efficiency on representative robot control tasks. Representative speedups: 3.3$\times$ on G1 Flip, 8.4$\times$ on G1 Walk Flat, and 11.0$\times$ on G1 Motion Tracking.}
    \label{fig:end_to_end_training}
    \vspace{-18pt}
\end{figure*}

Figure~\ref{fig:end_to_end_training} spans humanoid, motion-tracking, and dexterous-in-hand manipulation tasks and follows the progression from PPO to APPO and replay-based FastSAC/FlashSAC, indicating that the gain comes from learner--collector decoupling rather than a single task or algorithm configuration.

\begin{figure}[t]
    \centering
    \includegraphics[width=\linewidth]{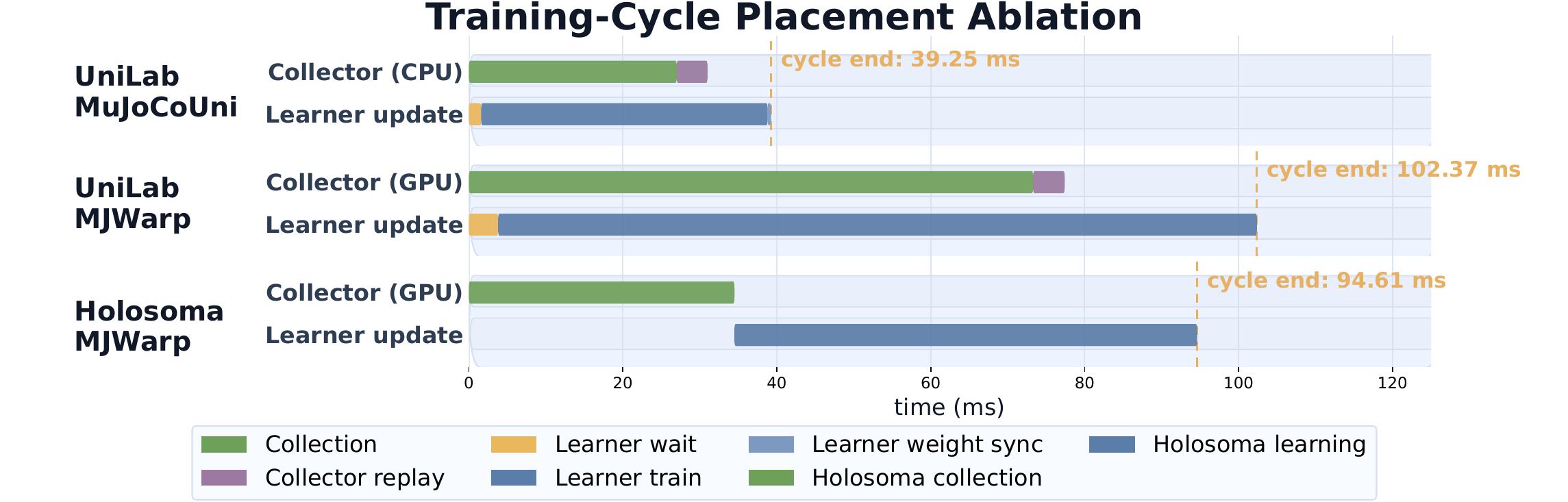}
    \caption{Training-cycle placement ablation. Holosoma is the FastSAC codebase used here, and MjWarp is its MuJoCo Warp backend. The figure compares where simulation collection and learning are placed during one learner cycle.}
    \label{fig:training_cycle_ablation}
    \vspace{-10pt}
\end{figure}

To further explain this gain, we add a learner-cycle ablation. Holosoma is the FastSAC codebase used here, and MjWarp denotes its MuJoCo Warp backend \citep{mujoco_warp}; Figure~\ref{fig:training_cycle_ablation} separates heterogeneous placement from runtime engineering alone: UniLab-MuJoCoUni completes collector work before the learner update ends, while attaching the same runtime to MjWarp lengthens the cycle because collector-side GPU simulation and learner updates share the same accelerator and contend for resources.

\begin{figure*}[t]
    \centering
    \includegraphics[width=\textwidth]{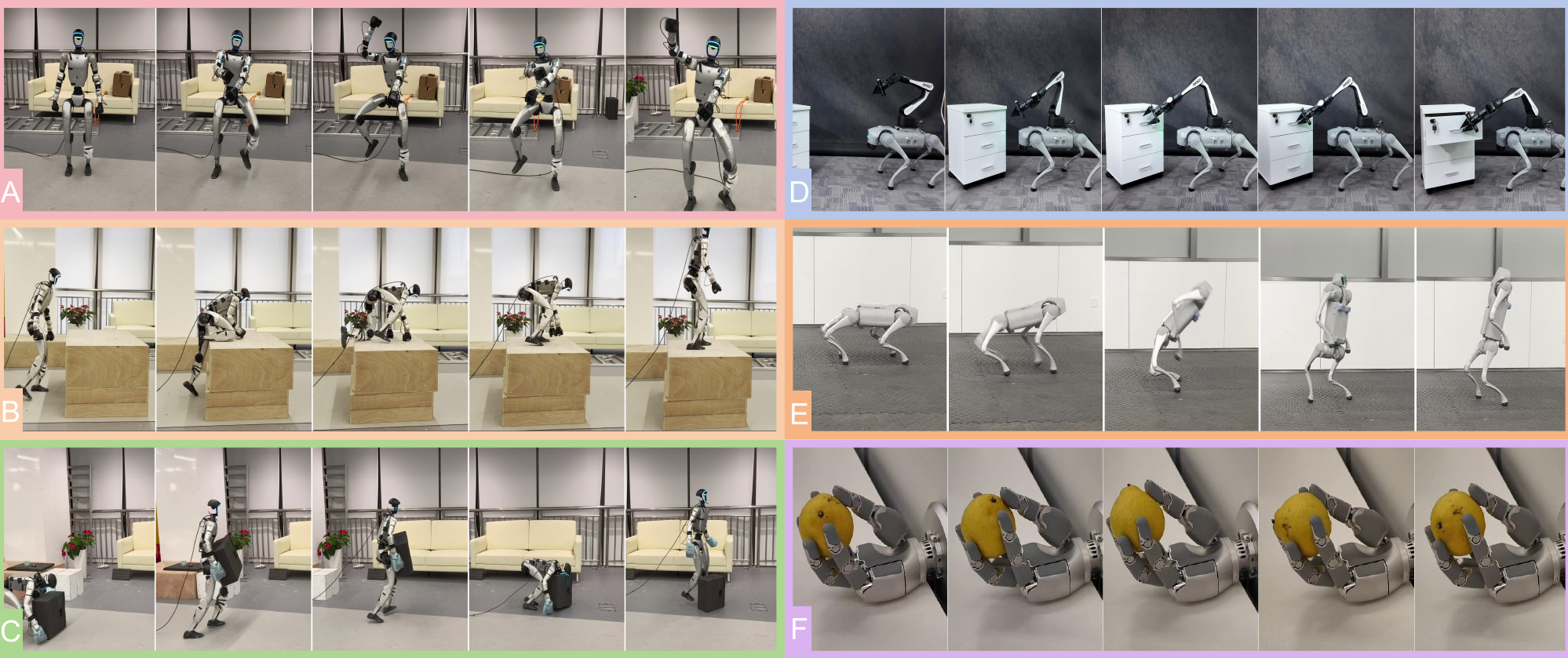}
    \caption{To-real experiment overview across six real-robot tasks.}
    \vspace{-16pt}
    \label{fig:to_real_experiments}
\end{figure*}

Figure~\ref{fig:to_real_experiments} gives an overview of the six to-real experiments and complements the end-to-end simulation results with deployment-side coverage.

\subsection{Dexterous In-Hand Rotation as a Systems Stress Test}
\label{subsec:dexterous_inhand}

SharpaWaveHand in-hand rotation adds contact-rich evidence beyond locomotion and motion tracking. The baseline starts from the public Sharpa-rl-lab PPO recipe on Isaac Lab, with object-scale randomization adjusted to match \method{}; it uses a fixed gravity direction with a built-in curriculum, whereas \method{} trains directly under randomized gravity directions without curriculum. In this task, the CPU MuJoCo version trains better, and \method{} reaches stronger HORA teacher policies within a shorter wall-clock budget; under the same-number friction setting, \method{}-SAC still reaches 1000+ reward in comparable time. The task uses a 22-DOF tactile hand to rotate a randomized free object and shows that \method{} supports dense simulation, stable learning, and different synchronization constraints in dexterous teacher training.

\subsection{Cross-Platform Evidence}
\label{subsec:cross_platform}

Finally, we report Apple macOS, AMD ROCm, and Intel XPU results to show practical trainability outside a single CUDA-centric setup, without claiming absolute throughput parity with the main Linux/CUDA workstation. Table~\ref{tab:platform_walltime} summarizes wall-clock training time across representative devices and tasks. Cross-platform execution is a practical consequence of the \method{} interface design.

\begin{figure*}[t]
    \centering
    \includegraphics[width=\textwidth]{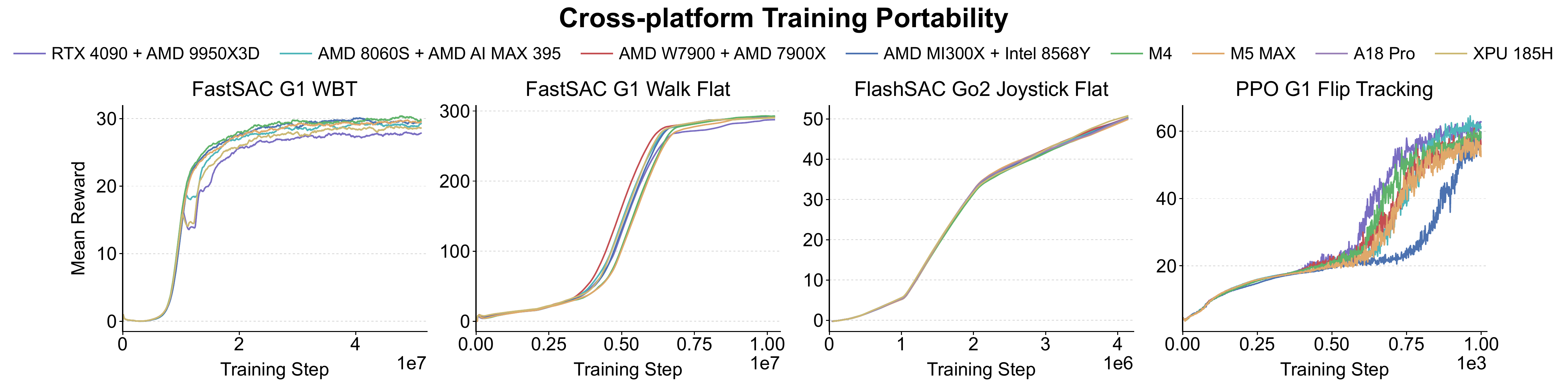}
    \caption{Cross-platform training overview on representative devices. The figure shows training curves and final performance on different platforms, complementing Table~\ref{tab:platform_walltime}.}
    \label{fig:platform_reward}
    \vspace{-14pt}
\end{figure*}

\begin{table}[t]
\centering
\caption{Wall-clock training time (min.).}
\label{tab:platform_walltime}
\scriptsize
\setlength{\tabcolsep}{3.5pt}
\renewcommand{\arraystretch}{0.9}
\begin{tabular}{@{}lcccc@{}}
\toprule
Device & FastSAC/G1 WBT & FastSAC/G1 Walk & FlashSAC/Go2 Joy. & PPO/G1 Flip \\
\midrule
RTX 4090 (Baseline)        & 58.8 & 18.3 & 6.0  & 109.0 \\
RTX 4090 + AMD 9950X3D     & 18.5 & 3.0  & 1.1  & 16.4 \\
AMD 8060S + AMD AI MAX 395 & 33.6 & 9.4  & 4.2  & 19.6 \\
M5 Max                     & 75.0 & 18.8 & 4.5  & 16.8 \\
\bottomrule
\end{tabular}
\vspace{-8pt}
\end{table}

\section{Conclusion}
\label{sec:conclusion}

This paper presented \method{}, a heterogeneous CPU-simulation / GPU-learning architecture for robot RL. By coordinating data movement, buffering, and synchronization through a unified runtime, \method{} improves end-to-end training efficiency by 3--10$\times$ across multiple robot embodiments, control workloads, and practical algorithms, while reducing dependence on the NVIDIA CUDA-based software stack and supporting Apple macOS, AMD ROCm, and Intel XPU backends. These results show that efficient training depends on high-throughput, well-coordinated simulation-learning execution, rather than requiring physics to reside on the GPU; \method{} therefore provides a systems counterexample showing that the design space for efficient training is broader than the current GPU-centric default suggests.

\section{Discussion}
\label{sec:discussion}

Our claim is not that GPU-resident simulation is obsolete. GPU simulation may remain preferable when simulator throughput is no longer the bottleneck or when larger accelerator-rich configurations are a better fit. \method{} broadens the design space for simulation-dominated robot control.

The speed of a GPU-centric stack comes from two coupled designs: simulation, rollout collection, and learning share a low-overhead execution path, while the physics backend is organized as GPU-friendly parallel computation. The former is a training-system organization principle; the latter is one hardware path for realizing it. This path is effective for regular, dense, and statically shaped computation, but dynamic contacts, sparse interactions, collision handling, and constraint solving can increase backend engineering pressure and make implementations depend more on specialization, buffer tuning, and static-allocation assumptions. Thus, this paper does not challenge the value of GPU simulators; it challenges the necessity claim that efficient robot RL training must use GPU-resident physics.

\section{Limitations}
\label{sec:limitations}

The main limitations follow from three assumptions. First, \method{} is most advantageous when training is simulation-dominated and simulation can be meaningfully decoupled from learning; on strictly synchronized pipelines or vision-based workloads dominated by rendering, perception, and representation learning, CPU/GPU decoupling may not hide the dominant cost and may therefore yield smaller gains. Second, our claim concerns end-to-end training efficiency in a controlled single-CPU/single-GPU setting, not absolute peak throughput at extreme scale; multi-GPU or larger distributed configurations may change the bottleneck and the hardware-allocation tradeoff. Third, the current implementation focuses on rigid-body robot control rather than deformable objects, soft bodies, or fluids. Future work should extend the same runtime analysis to vision-dominated tasks, larger systems, and non-rigid physics to identify where the heterogeneous design fails and what scheduling, communication, or backend changes are needed.

\acknowledgments{
We thank Apple and AMD for providing hardware platforms for development and evaluation, and for assisting with platform adaptation.
We are also sincerely grateful to the mjlab team for open-sourcing their excellent work, whose engineering practices provided valuable reference for this project.
We also thank early users of \method{} and the students in Tsinghua University's Spring 2026 Deep Reinforcement Learning course for their use and feedback.
}

\bibliography{references}

\clearpage
\input{Sections/8_Appendix}

\end{document}